\documentclass[11pt]{article}
\usepackage{geometry}
\usepackage{authblk}
\usepackage{coling2020}
\usepackage{times}
\usepackage{url}
\usepackage{latexsym}
\usepackage{microtype}
\usepackage{graphicx}

\colingfinalcopy % Uncomment this line for all SemEval submissions

\title{UTMN at SemEval-2020 Task 11: A Kitchen Solution to Automatic Propaganda Detection}

\author{Elena Mikhalkova, Nadezhda Ganzherli, Anna Glazkova, Yuliya Bidulya\\
    Tyumen State University \\
    Tyumen, Russia \\
    {\tt e.v.mikhalkova@utmn.ru}}

\date{}

\begin{document}
\maketitle
\begin{abstract}
  The article describes a fast solution to propaganda detection at SemEval-2020 Task 11, based on feature adjustment. We use per-token vectorization of features and a simple Logistic Regression classifier to quickly test different hypotheses about our data. We come up with what seems to us the best solution, however, we are unable to align it with the result of the metric suggested by the organizers of the task. We test how our system handles class and feature imbalance by varying the number of samples of two classes (Propaganda and None) in the training set, the size of a context window in which a token is vectorized and combination of vectorization means. The result of our system at SemEval2020 Task 11 is F-score=0.37.
\end{abstract}

\section{Introduction}
\label{intro}

\blfootnote{

    \hspace{-0.65cm}  % space normally used by the marker
    This work is licensed under a Creative Commons 
    Attribution 4.0 International License.
    License details:
    \url{http://creativecommons.org/licenses/by/4.0/}.
}

Propaganda is a complex phenomenon that was studied in psychology \cite{doob1935propaganda}, sociology~\cite{jackall1995propaganda,klaehn2010propaganda}, theory of communication \cite{jowett2018propaganda,severin1997communication}, pedagogy \cite{hobbs2014teaching,smith1974critically}, history~\cite{taylor2013munitions}, linguistics~\cite{lukin2013journalism}, and other sciences. Naturally, the task of automatic propaganda detection has been set and approached in different ways. \textit{Proppy}, an on-line service, detects propaganda in news articles and clusters them according to an index of propaganda~\cite{barron2019proppy}. It daily analyzes emerging texts, identifies events described in them, discards near-duplicates, and, lastly, computes propaganda index on the basis of \textit{n}-gram features, vocabulary, its richness, style, readability, and NEws LAndscape (NELA)~\cite{horne2018sampling}. In the Propaganda Analysis Project, the propaganda detection is more focused on locating propaganda within a text. At the Hack News Datathon~\footnote{\url{https://www.datasciencesociety.net/hack-news-datathon/}} Task 3 was to detect a fragment containing propaganda. However, it was paired with classifying it according to 18 techniques~\footnote{These 18 techniques have grown from the 1930s American Institute of Propaganda Analysis materials~\cite{miller1939propaganda} and more recent investigations both into the tools of propaganda \cite[and others]{torok2015symbiotic,teninbaum2009reductio}, and into the rules of good argument \cite{weston2018rulebook}.}, same as the Fragment Level Classification at the shared task of ``Fine-Grained Propaganda Detection'' (EMNLP2019)~\cite{EMNLP19DaSanMartino}. The competition of Da San Martino et al.~\shortcite{DaSanMartinoSemeval20task11} evaluates these two stages separately. Although it is hard to map results of previous competitions on the current one, the former revealed some life hacks that can be of use in the latter. For example, as noted by H.T. Madabushi~\shortcite{madabushi2019cost}, extracting fragments of propaganda is similar to the task of Named Entity Recognition, in that they are both span extraction tasks. Their system ProperGander was partially based on the BERT~\cite{devlin-etal-2019-bert} solution for NER. However, probably, the main lesson learned from the first two events was about the necessity of using BERT (or at least a deep-learning neural network) to achieve a state-of-the-art result.

In the current article, we proudly present a no-BERT and even no-deep-learning solution for the task of Span Identification. Our choice not to utilize contextualized vectors such as BERT is grounded by the following reasons. (1) In the annotation to ``Poor Man's BERT: Smaller and Faster Transformer Models''~\cite{poorman}, it is noted that BERT-based models require certain laboratory facilities, and our team is not from a lab. (2) Even in a lab, BERT takes time to train. So, it is also a slow man's choice that deprives one from quickly testing many hypotheses. (3) We were unable to find a neural classifier that would replace a simpler classifier in our best solution before the end of the competition. Even a pre-trained BERT model of T.Wolf et al.~\shortcite{Wolf2019HuggingFacesTS} takes about 40 minutes per 1 epoch to train in Google Colab, when our model takes 2-3 minutes to train and predict.

The paper is organized as follows: we first describe our idea of class and feature imbalance in propaganda; we, then, outline the basics of our classifier; we devote a paragraph to testing hypotheses about our data and conclude about the usability of our system.

\section{Propaganda in News Articles}

The \textbf{verbal propaganda in a news article} is a stretch of text written with an intent to influence the reader's opinion~\footnote{In this respect Edward Sapir's conclusions hold true: ``It is generally difficult to make a complete divorce between objective reality and our linguistic symbols of reference to it; and things, qualities and events are on the whole felt to be what they are called.''~\cite{sapir1933language}. But at the same time, what you consider propaganda depends on your point of view, cf. Edward Bernays's opinion as cited by Hobbs et al.~\shortcite{hobbs2014teaching}: ``The advocacy of what we believe in is education. The advocacy of what we don't believe is propaganda''.} with arguments that lead to conclusions advantageous to the author. The communicative intent is to control and manipulate the reader. The tools are cognitive, logical, emotional and other means to confuse the reader (verbal fouling), and the verbal representation adds up information to what has to be said as a fact in a news article as a genre: the known facts and the believed-to-be truth~\footnote{``Ellul believed that truth does not separate propaganda from ``moral forms'' because propaganda uses truth, half-truth, and limited truth. A similar statement from British Labour politician Richard Crossman is that ``the art of propaganda is not telling lies but rather seeing the truth you require and giving it mixed up with some truths the audience wants to hear'' (Higham, 2013, p.2).''~\cite{jowett2018propaganda}}. Hence, we expect that propaganda enlarges the size of a textual unit where it appears. This unit is usually a sentence: according to Da San Martino et al.~\shortcite{EMNLP19DaSanMartino}, propaganda takes about half a sentence. And the sentences with propaganda tend to be longer, as reported by creators of the PIG system at the Hack News Datathon~\footnote{{https://www.datasciencesociety.net/datathon-hacknews-solution-pig-propaganda-identification-group/}}.

The attribute of secrecy requires that propaganda looks like its non-propagandist context. So, it should not differ much from the context in its lexical, grammatical or stylistic expression: there should be a difference, but slight. Rashkin et al.~\shortcite{rashkin2017truth} look at four kinds of news articles (reliable, hoax, propagandistic and satiric) and find that they have linguistic features that are more frequent in fake news (the difference is statistically significant): pronouns `I' and `you' and their forms, modal, action, manner adverbs, words semantically related to {\em swear, sexual, see, negation}~\footnote{According to LIWC lexicon~\cite{pennebaker2015development}.}, strong and weak subjective etc.~\footnote{There seem to be reasonable grounds behind these features. For example, it is easier to negate something that has been never or rarely encountered before, as the burden of proof lies on somebody who rather states existence, cf. the historical case of in-existence of black swans.} At the same time they enlist some features which are \textbf{not} in the fake news, but these features are fewer. As mentioned, propaganda should look like news, but with a bit of additional ``foul'' content. Hence, the method of semantic vectors (with due regard for other methods) must be useful in propaganda detection to extract semantic features of the class ``propaganda'', but it will not be as useful in defining the class of ``non-propaganda''.

To sum up, there are linguistic features that are more common in propaganda than in non-propaganda, and there are nearly no linguistic features of non-propaganda that are uncommon in propaganda ({\em feature imbalance}). Further, propaganda is less common than non-propaganda ({\em class imbalance}). Finally, propaganda takes about half a sentence; sentences with propaganda tend to be longer; propaganda is emotionally colored.

\section{System Architecture}
\label{sysarch}

Detection of propaganda as a fragment of an article (fragment or span identification) presupposes that the minimum textual unit that holds it is a word form. However, if it neighbors with punctuation marks, numbers, etc. the latter can also be attributed to the fragment. Hence, the per-token approach is very common in solutions to tasks organized by the Propaganda Analysis Project: Yoosuf  and Yang~\shortcite{yoosuf-yang-2019-fine} use BERT to classify each token into 20 classes (18 propaganda techniques, none and ``auxiliary''); Gupta et al.~\shortcite{gupta-etal-2019-neural} create a token vector from a combination of features, including an embedding vector of a word and sentence; Ek and Ghanimifard~\shortcite{ek-ghanimifard-2019-synthetic} combine a vector of a token from three vectors of contextualized embeddings (ELMo~\cite{Peters:2018}, BERT~\cite{devlin-etal-2019-bert}, GROVER~\cite{DBLP:journals/corr/abs-1905-12616}); Alhindi et al.~\shortcite{alhindi-etal-2019-fine} and Yoosuf and Yang~\shortcite{yoosuf-yang-2019-fine} use not only word but also character embeddings, as some morphemes, e.g. ``-ist'', can be frequent in propaganda.

As mentioned, in this project we decided to focus on a fast solution that would allow us to test many hypotheses about our data. As spans should not contain just words, we chose a token as a minimum unit of a span, regardless whether it is a word form, a punctuation mark, etc. With tokens that are not actual words, it is important to rely on the context of each token --- this is what we base our model on. A \textbf{context window} is a chunk of text that is within {\em n} tokens to the left and {\em n} tokens to the right from the given token; the adjustable parameter of the context window is its size {\em n}. If the token has too few tokens to the left or right, the context window simply shrinks. We tokenize texts with SpaCy~\cite{honnibal-johnson:2015:EMNLP}: the EnCoreWebLg model~\footnote{\url{https://spacy.io/models/en#en_core_web_lg}. Note that when Spacy detects a pronoun, it adds '-PRON-' to the list of lemmas.}. The machine learning algorithm that we finally chose is the Logistic Regression (LR) implemented in Scikit-Learn~\cite{scikit-learn}. The reasons for that were, first, the fastness of LR and, second, it proved to be one of the most efficient classifiers for the data models that we tested at the preliminary stage of research. Also, LR can be considered a shallow neural network and, hence, a baseline to test how neural networks treat the data model.

\subsection{Data Model, Adjustable Parameters, Evaluation}

As mentioned, the unit of classification is a token extracted with SpaCy. The token is considered within a context window of length {\em n} tokens to the right and left from the given token. Our data model is grounded by our approach. For each token it combines:

\begin{enumerate}
    \item An embedding vector of the {\em token}: size 200, window = 7, acquired with Word2vec model of Gensim~\cite{rehurek_lrec}. This is a safety model in case SpaCy does not know the token. The model is trained on the sentences of all articles in the training, development and test sets. Prepocessing, lemmatization and sentence splitting are done with SpaCy~\footnote{However, SpaCy sometimes lemmatizes the same words differently. So, some few tokens that are not in Word2vec are dropped from classification. The older the Word2vec model is, the more often such tokens occur.}.
    \item An embedding vector of the {\em token}: size 300, acquired with SpaCy ``vector'' command.
    \item An embedding vector of the {\em context window}: size 300, acquired with SpaCy ``vector'' command. To get it and the two next vectors, we cut out the part of the text that belongs to the context window and only then vectorize it.
    \item A sentiment vector of the {\em context window}: size 4, acquired with NLTK Vader Sentiment Intensity Analyzer~\cite{Hutto2014VADERAP,bird2009natural}.
    \item A semantic vector of the {\em context window} of size 39, acquired with our vectorizer based on Roget's Thesaurus ~\cite{mikhalkova2017detecting,mikhalkova2017punfields}.
\end{enumerate}

The final length of the vector is 843. At the competition, we normalized the vectors  with Scikit-Learn ``preprocessing.normalize''.

We do not add such a parameter as the length of the sentence in which the token is found (although we mentioned it as a characteristic one) as, when added to the vector, even normalized, it significantly decreases result of the Logistic Regression.

Given the feature- and class-imbalance, we have the following parameters to tune in our data model:

\begin{enumerate}
    \item Size of context window {\em n}.
    \item Number of propaganda and non-propaganda samples in the training set.
    \item Number of samples of each type of propaganda in the class of propaganda in the training set.
    \item Number of articles to give non-propaganda samples for the training set.
\end{enumerate}

\begin{figure}
    \centering
    \includegraphics[scale=0.4]{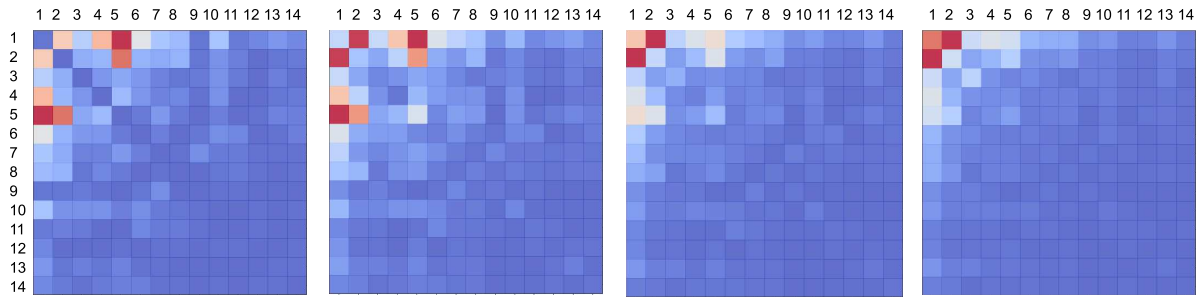}
    \caption{Overlapping and neighboring fragments of propaganda calculated per character: blue -- fewer overlaps, neighbours; red -- otherwise. Square 1: full overlaps of characters; Squares 2-4: full overlaps plus neighbours within 5, 20, 30 characters to the left and right. Classes: 1 - Loaded L., 2 - Name C., 3 - Repetition, 4 - Exagg., 5 - Doubt, 6 - Appeal to Fear, 7 - Flag-W., 8 - Causal O., 9 - Slogans, 10 - Appeal to A., 11 - Black-and-White C., 12 - Thought-Term., 13 - Whatabout., 14 Bandwagon.}
    \label{fig:Heatmap}
\end{figure}

As concerns probability of the same tokens belonging to more than one fragment of propaganda (due to difference in classes), we do not consider it in our model. The heat-map on Figure~\ref{fig:Heatmap}, the left-most square, shows that propaganda fragments do not co-occur so often: most of the field is blue. This co-occurrence seems to depend on how frequent the class is (see class ``Loaded Language'', for example) with one exception: class ``Doubt'', although it is not among the most frequent, tends to co-occur with Loaded Language and Name Calling more often than its neighbors. When we enlarge the interval including not only overlaps, but neighboring within $n$ characters (the three other squares), this trend holds.

To test our algorithm before the competition we applied F-score for per-token classification: true positive values are \textbf{tokens} correctly classified as propaganda and false negative are \textbf{tokens} correctly classified as non-propaganda. During the competition, we had difficulty aligning our results with the results we got for the development set in the system offered by the organizers. The organizers' system is per-character F-score that also takes into account the number of fragments (spans). For now, it is hard to say whether our results are hard to align due to per-token and per-character approach or that the organizers' approach punishes for too few or too many fragments. In this article, to evaluate the result we will simply be using percentage of correctly and wrongly classified \textbf{tokens} for the two classes: Propaganda and None.

\subsection{Testing}

\begin{figure}
    \centering
    \includegraphics[scale=0.7]{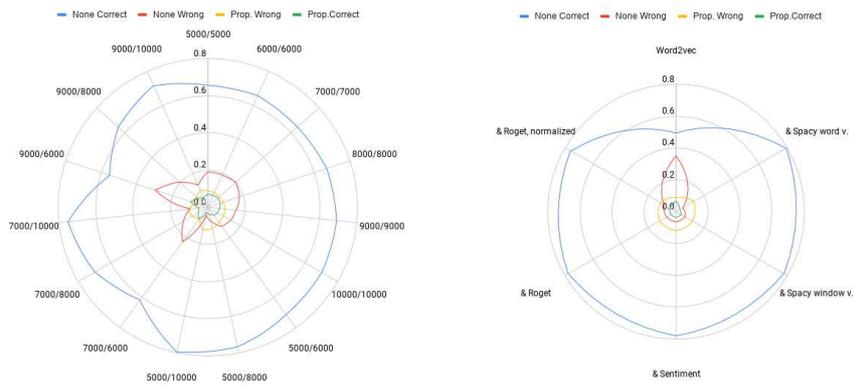}
    \caption{Left: system's result on combinations of features ``Propaganda'' and ``None''. Right (clockwise): system's result when we add up a sub-vector to the result vector of a token. Y-axis: percentage of tokens correctly and wrongly classified in each class.}
    \label{fig:Radar}
\end{figure}

To tune our model, we experimented with some of the adjustable parameters mentioned above (each parameter is a hypothesis about our data): the number of features of each class and the size of the context window. The ML classifier is LogisticRegression(solver='liblinear',penalty='l2',C=0.1). Given two sets of token vectors: 10,000 from fragments with propaganda and 10,000 --- without propaganda~\footnote{Although it would be easier to take the first 10,000 non-propagandist tokens, tests showed that the more articles give None samples, the better.} of the competition's official train set, and the vectors of test set, let us first check how the number of vectors of each class influences the mentioned percentages. The radar chart on Figure~\ref{fig:Radar}, left, shows that an equal number of vectors of each class (Propaganda and None) in the train set gives an even performance, but when we create disbalance, e.g. put 7,000 vectors of class Propaganda and 10,000 of None in the train set, it can lead to a more favorable treatment of a class. For further testing we will stick with the combination of 7,000 Propaganda and 10,000 None as we chose this combination at the competition. It gives a very good performance for the class None and a fairly good one for Propaganda.

\begin{figure}
    \centering
    \includegraphics[scale=0.9]{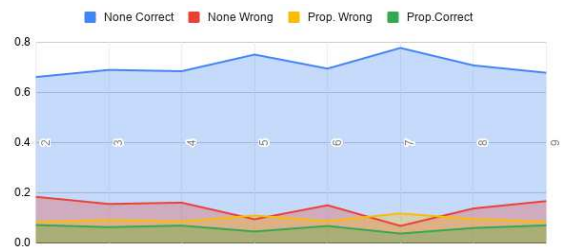}
    \caption{Percentage of tokens correctly and wrongly classified in each context window. X-axis: the size of the context window.}
    \label{fig:CW}
\end{figure}

Now, let us test the size of the window. Figure~\ref{fig:CW} shows that there are two peaks where classification of None is best: 5 and 7. And, although correct classification of Propaganda is lower than usual at them, it is quite even compared to performance at other context windows. For the competition we have chosen the context window of size 7.

As for inclusion of different vector types in the final vector, the radar chart in Figure~\ref{fig:Radar}, right, shows how adding these types (if we go clockwise starting with just Word2vec vector) balances the proportion of False Positive and True Negative cases, and increases the number of correctly classified Propaganda tokens. However, normalizing the final vector now seems to us wrong, as it tangibly decreases the number of correctly classified None vectors.

Our official result at SemEval-2020, Task 11 is: F-score=0.37490, precision=0.37967, recall=0.37025. Our code is available at \url{https://github.com/evrog/Propaganda-Detection-at-SemEval-2020}.

\section{Conclusion}

We have described our system of automatic propaganda detection at SemEval-2020. The system is designed so as to allow us to quickly test many feature parameters. We call it a kitchen solution in the title: coming in small, but handy and utilizable blocks. In Google Colab, our classifier takes 141 seconds (without accelerator) and 128 seconds (with GPU accelerator) to train and calculate the result on the test data.

The idea of feature imbalance has led us to using an unbalanced combination of Propaganda and None token vectors in the data set. Also, we have experimented with the size of a context window in which propaganda becomes most discernible and ended up with size 7, the number which peculiarly correlates with data on human short-term memory: ``the 7-item limit''~\cite{coon2008introduction}. It seems to us that this correlation is suggestive of the idea that, in the end, what we, as readers, are able to call propaganda, is propaganda.

\bibliographystyle{coling}
%\bibliography{semeval2020}
\bibliography{Mikhalkova}

\begin{thebibliography}{}

\bibitem[\protect\citename{Alhindi \bgroup et al.\egroup
  }2019]{alhindi-etal-2019-fine}
Tariq Alhindi, Jonas Pfeiffer, and Smaranda Muresan.
\newblock 2019.
\newblock Fine-tuned neural models for propaganda detection at the sentence and
  fragment levels.
\newblock In {\em Proceedings of the Second Workshop on Natural Language
  Processing for Internet Freedom: Censorship, Disinformation, and Propaganda},
  pages 98--102, Hong Kong, China, nov. Association for Computational
  Linguistics.

\bibitem[\protect\citename{Barr{\'o}n-Cedeno \bgroup et al.\egroup
  }2019]{barron2019proppy}
Alberto Barr{\'o}n-Cedeno, Giovanni Da~San~Martino, Israa Jaradat, and Preslav
  Nakov.
\newblock 2019.
\newblock Proppy: A system to unmask propaganda in online news.
\newblock In {\em Proceedings of the AAAI Conference on Artificial
  Intelligence}, volume~33, pages 9847--9848.

\bibitem[\protect\citename{Bird \bgroup et al.\egroup }2009]{bird2009natural}
Steven Bird, Ewan Klein, and Edward Loper.
\newblock 2009.
\newblock {\em Natural language processing with Python: analyzing text with the
  natural language toolkit}.
\newblock O'Reilly Media, Inc.

\bibitem[\protect\citename{Childs}1936]{doob1935propaganda}
Harwood~L. Childs.
\newblock 1936.
\newblock {\em The American Political Science Review}, 30(2):389--390.

\bibitem[\protect\citename{Coon and Mitterer}2008]{coon2008introduction}
D.~Coon and J.O. Mitterer.
\newblock 2008.
\newblock {\em Introduction to Psychology: Gateways to Mind and Behavior}.
\newblock Available Titles CengageNOW Series. Cengage Learning.

\bibitem[\protect\citename{Da~San~Martino \bgroup et al.\egroup
  }2019]{EMNLP19DaSanMartino}
Giovanni Da~San~Martino, Seunghak Yu, Alberto Barr\'{o}n-Cede\~no, Rostislav
  Petrov, and Preslav Nakov.
\newblock 2019.
\newblock Fine-grained analysis of propaganda in news articles.
\newblock In {\em Proceedings of the 2019 Conference on Empirical Methods in
  Natural Language Processing and the 9th International Joint Conference on
  Natural Language Processing, EMNLP-IJCNLP 2019}, EMNLP-IJCNLP 2019, Hong
  Kong, China, November.

\bibitem[\protect\citename{Da~San~Martino \bgroup et al.\egroup
  }2020]{DaSanMartinoSemeval20task11}
Giovanni Da~San~Martino, Alberto Barr\'{o}n-Cede\~no, Henning Wachsmuth,
  Rostislav Petrov, and Preslav Nakov.
\newblock 2020.
\newblock {SemEval}-2020 task 11: Detection of propaganda techniques in news
  articles.
\newblock In {\em Proceedings of the 14th International Workshop on Semantic
  Evaluation}, SemEval 2020, Barcelona, Spain, September.

\bibitem[\protect\citename{Devlin \bgroup et al.\egroup
  }2019]{devlin-etal-2019-bert}
Jacob Devlin, Ming-Wei Chang, Kenton Lee, and Kristina Toutanova.
\newblock 2019.
\newblock {BERT}: Pre-training of deep bidirectional transformers for language
  understanding.
\newblock In {\em Proceedings of the 2019 Conference of the North {A}merican
  Chapter of the Association for Computational Linguistics: Human Language
  Technologies, Volume 1 (Long and Short Papers)}, pages 4171--4186,
  Minneapolis, Minnesota, June. Association for Computational Linguistics.

\bibitem[\protect\citename{Ek and
  Ghanimifard}2019]{ek-ghanimifard-2019-synthetic}
Adam Ek and Mehdi Ghanimifard.
\newblock 2019.
\newblock Synthetic propaganda embeddings to train a linear projection.
\newblock In {\em Proceedings of the Second Workshop on Natural Language
  Processing for Internet Freedom: Censorship, Disinformation, and Propaganda},
  pages 155--161, Hong Kong, China, November. Association for Computational
  Linguistics.

\bibitem[\protect\citename{Gupta \bgroup et al.\egroup
  }2019]{gupta-etal-2019-neural}
Pankaj Gupta, Khushbu Saxena, Usama Yaseen, Thomas Runkler, and Hinrich
  Sch{\"u}tze.
\newblock 2019.
\newblock Neural architectures for fine-grained propaganda detection in news.
\newblock In {\em Proceedings of the Second Workshop on Natural Language
  Processing for Internet Freedom: Censorship, Disinformation, and Propaganda},
  pages 92--97, Hong Kong, China, November. Association for Computational
  Linguistics.

\bibitem[\protect\citename{Hobbs and McGee}2014]{hobbs2014teaching}
Renee Hobbs and Sandra McGee.
\newblock 2014.
\newblock Teaching about propaganda: An examination of the historical roots of
  media literacy.
\newblock {\em Journal of Media Literacy Education}, 6(2):5.

\bibitem[\protect\citename{Honnibal and
  Johnson}2015]{honnibal-johnson:2015:EMNLP}
Matthew Honnibal and Mark Johnson.
\newblock 2015.
\newblock An improved non-monotonic transition system for dependency parsing.
\newblock In {\em Proceedings of the 2015 Conference on Empirical Methods in
  Natural Language Processing}, pages 1373--1378, Lisbon, Portugal, September.
  Association for Computational Linguistics.

\bibitem[\protect\citename{Horne \bgroup et al.\egroup
  }2018]{horne2018sampling}
Benjamin~D Horne, Sara Khedr, and Sibel Adali.
\newblock 2018.
\newblock Sampling the news producers: A large news and feature data set for
  the study of the complex media landscape.
\newblock In {\em Twelfth International AAAI Conference on Web and Social
  Media}.

\bibitem[\protect\citename{Hutto and Gilbert}2014]{Hutto2014VADERAP}
Clayton~J. Hutto and Eric Gilbert.
\newblock 2014.
\newblock Vader: A parsimonious rule-based model for sentiment analysis of
  social media text.
\newblock In {\em ICWSM}.

\bibitem[\protect\citename{Jackall}1995]{jackall1995propaganda}
Robert Jackall.
\newblock 1995.
\newblock {\em Propaganda}, volume~8.
\newblock NYU Press.

\bibitem[\protect\citename{Jowett and O'Donnell}2018]{jowett2018propaganda}
Garth~S Jowett and Victoria O'Donnell.
\newblock 2018.
\newblock {\em Propaganda \& persuasion}.
\newblock Sage Publications.

\bibitem[\protect\citename{Klaehn and Mullen}2010]{klaehn2010propaganda}
Jeffery Klaehn and Andrew Mullen.
\newblock 2010.
\newblock The propaganda model and sociology: understanding the media and
  society.
\newblock {\em Synaesthesia: Communication Across Cultures}, 1(1):10--23.

\bibitem[\protect\citename{Lukin}2013]{lukin2013journalism}
Annabelle Lukin.
\newblock 2013.
\newblock Journalism, ideology and linguistics: The paradox of chomsky’s
  linguistic legacy and his ‘propaganda model’.
\newblock {\em Journalism}, 14(1):96--110.

\bibitem[\protect\citename{Madabushi \bgroup et al.\egroup
  }2019]{madabushi2019cost}
Harish~Tayyar Madabushi, Elena Kochkina, and Michael Castelle.
\newblock 2019.
\newblock Cost-sensitive bert for generalisable sentence classification with
  imbalanced data.
\newblock {\em EMNLP-IJCNLP 2019}, pages 125--134.

\bibitem[\protect\citename{Mikhalkova and
  Karyakin}2017a]{mikhalkova2017detecting}
Elena Mikhalkova and Yuri Karyakin.
\newblock 2017a.
\newblock Detecting intentional lexical ambiguity in english puns.
\newblock In {\em Proceedings of the International Conference "Dialogue 2017"
  Moscow, May 31-June 3, 2017}.

\bibitem[\protect\citename{Mikhalkova and
  Karyakin}2017b]{mikhalkova2017punfields}
Elena Mikhalkova and Yuri Karyakin.
\newblock 2017b.
\newblock Punfields at semeval-2017 task 7: Employing roget’s thesaurus in
  automatic pun recognition and interpretation.
\newblock In {\em Proceedings of the 11th International Workshop on Semantic
  Evaluation (SemEval-2017)}, pages 426--431.

\bibitem[\protect\citename{Miller}1939]{miller1939propaganda}
Clyde~R Miller.
\newblock 1939.
\newblock Propaganda and the european war.
\newblock {\em The Clearing House: A Journal of Educational Strategies, Issues
  and Ideas}, 14(2):67--73.

\bibitem[\protect\citename{Pedregosa \bgroup et al.\egroup }2011]{scikit-learn}
F.~Pedregosa, G.~Varoquaux, A.~Gramfort, V.~Michel, B.~Thirion, O.~Grisel,
  M.~Blondel, P.~Prettenhofer, R.~Weiss, V.~Dubourg, J.~Vanderplas, A.~Passos,
  D.~Cournapeau, M.~Brucher, M.~Perrot, and E.~Duchesnay.
\newblock 2011.
\newblock Scikit-learn: Machine learning in {P}ython.
\newblock {\em Journal of Machine Learning Research}, 12:2825--2830.

\bibitem[\protect\citename{Pennebaker \bgroup et al.\egroup
  }2015]{pennebaker2015development}
James~W Pennebaker, Ryan~L Boyd, Kayla Jordan, and Kate Blackburn.
\newblock 2015.
\newblock The development and psychometric properties of liwc2015.
\newblock Technical report, Austin, TX: University of Texas at Austin.

\bibitem[\protect\citename{Peters \bgroup et al.\egroup }2018]{Peters:2018}
Matthew~E. Peters, Mark Neumann, Mohit Iyyer, Matt Gardner, Christopher Clark,
  Kenton Lee, and Luke Zettlemoyer.
\newblock 2018.
\newblock Deep contextualized word representations.
\newblock In {\em Proc. of NAACL}.

\bibitem[\protect\citename{Rashkin \bgroup et al.\egroup
  }2017]{rashkin2017truth}
Hannah Rashkin, Eunsol Choi, Jin~Yea Jang, Svitlana Volkova, and Yejin Choi.
\newblock 2017.
\newblock Truth of varying shades: Analyzing language in fake news and
  political fact-checking.
\newblock In {\em Proceedings of the 2017 Conference on Empirical Methods in
  Natural Language Processing}, pages 2931--2937.

\bibitem[\protect\citename{{\v R}eh{\r u}{\v r}ek and Sojka}2010]{rehurek_lrec}
Radim {\v R}eh{\r u}{\v r}ek and Petr Sojka.
\newblock 2010.
\newblock {Software Framework for Topic Modelling with Large Corpora}.
\newblock In {\em {Proceedings of the LREC 2010 Workshop on New Challenges for
  NLP Frameworks}}, pages 45--50, Valletta, Malta, May. ELRA.
\newblock \url{http://is.muni.cz/publication/884893/en}.

\bibitem[\protect\citename{Sajjad \bgroup et al.\egroup }2020]{poorman}
Hassan Sajjad, Fahim Dalvi, Nadir Durrani, and Preslav Nakov.
\newblock 2020.
\newblock Poor man's bert: Smaller and faster transformer models.

\bibitem[\protect\citename{Sapir}1933]{sapir1933language}
Edward Sapir.
\newblock 1933.
\newblock Language in: Encyclopaedia of the social sciences.
\newblock {\em New York}, 9:155--169.

\bibitem[\protect\citename{Severin \bgroup et al.\egroup
  }1997]{severin1997communication}
Werner~Joseph Severin, James~W Tankard, et~al.
\newblock 1997.
\newblock {\em Communication theories: Origins, methods, and uses in the mass
  media}.
\newblock Longman New York.

\bibitem[\protect\citename{Smith}1974]{smith1974critically}
Bonnie Smith.
\newblock 1974.
\newblock {\em Critically Reading for Propaganda Techniques in Grade Six.}
\newblock {Ph.D.} thesis, M. Ed. Thesis, Rutgers University, The State
  University of New Jersey.

\bibitem[\protect\citename{Taylor}2013]{taylor2013munitions}
Philip~M Taylor.
\newblock 2013.
\newblock {\em Munitions of the mind: A history of propaganda from the ancient
  world to the present era}.
\newblock Manchester University Press.

\bibitem[\protect\citename{Teninbaum}2009]{teninbaum2009reductio}
Gabriel~H Teninbaum.
\newblock 2009.
\newblock Reductio ad hitlerum: Trumping the judicial nazi card.
\newblock {\em Mich. St. L. Rev.}, page 541.

\bibitem[\protect\citename{Torok}2015]{torok2015symbiotic}
Robyn Torok.
\newblock 2015.
\newblock {\em Symbiotic radicalisation strategies: Propaganda tools and neuro
  linguistic programming}.
\newblock SRI Security Research Institute, Edith Cowan University, Perth,
  Western~….

\bibitem[\protect\citename{Weston}2018]{weston2018rulebook}
Anthony Weston.
\newblock 2018.
\newblock {\em A rulebook for arguments}.
\newblock Hackett Publishing.

\bibitem[\protect\citename{Wolf \bgroup et al.\egroup
  }2019]{Wolf2019HuggingFacesTS}
Thomas Wolf, Lysandre Debut, Victor Sanh, Julien Chaumond, Clement Delangue,
  Anthony Moi, Pierric Cistac, Tim Rault, R'emi Louf, Morgan Funtowicz, and
  Jamie Brew.
\newblock 2019.
\newblock Huggingface's transformers: State-of-the-art natural language
  processing.
\newblock {\em ArXiv}, abs/1910.03771.

\bibitem[\protect\citename{Yoosuf and Yang}2019]{yoosuf-yang-2019-fine}
Shehel Yoosuf and Yin Yang.
\newblock 2019.
\newblock Fine-grained propaganda detection with fine-tuned {BERT}.
\newblock In {\em Proceedings of the Second Workshop on Natural Language
  Processing for Internet Freedom: Censorship, Disinformation, and Propaganda},
  pages 87--91, Hong Kong, China, November. Association for Computational
  Linguistics.

\bibitem[\protect\citename{Zellers \bgroup et al.\egroup
  }2019]{DBLP:journals/corr/abs-1905-12616}
Rowan Zellers, Ari Holtzman, Hannah Rashkin, Yonatan Bisk, Ali Farhadi,
  Franziska Roesner, and Yejin Choi.
\newblock 2019.
\newblock Defending against neural fake news.
\newblock {\em CoRR}, abs/1905.12616.

\end{thebibliography}

\end{document}